\documentclass[lettersize,journal]{IEEEtran}
\usepackage{amsmath,amsfonts}
\usepackage{algorithmic}
\usepackage{algorithm}
\usepackage{array}
\usepackage[caption=false,font=normalsize,labelfont=sf,textfont=sf]{subfig}
\usepackage{textcomp}
\usepackage{stfloats}
\usepackage{url}
\usepackage{verbatim}
\usepackage{graphicx}
\usepackage{cite}
\usepackage{xcolor}
\usepackage{multirow}
\usepackage{caption}
\usepackage{arydshln}
\usepackage{hyperref}
\usepackage{booktabs}
\usepackage{array}
\begin{document}


\title{A Comprehensive Survey on Human Video Generation: Challenges, Methods, and Insights}

\author{Wentao Lei,
        Jinting Wang,
        Fengji Ma,
        Guanjie Huang,
        Li Liu
\IEEEcompsocitemizethanks{
\IEEEcompsocthanksitem All authors are with the Hong Kong University of Science and Technology (Guangzhou), Guangzhou 511458, China. \protect


}
\thanks{All authors are equal contributions.}
\thanks{Corresponding author: Li Liu, avrillliu@hkust-gz.edu.cn.}
}




\maketitle

\begin{abstract}
Human video generation is a dynamic and rapidly evolving task that aims to synthesize 2D human body video sequences with generative models given control conditions such as text, audio, and pose. With the potential for wide-ranging applications in film, gaming, and virtual communication, the ability to generate natural and realistic human video is critical. Recent advancements in generative models have laid a solid foundation for the growing interest in this area. Despite the significant progress, the task of human video generation remains challenging due to the consistency of characters, the complexity of human motion, and difficulties in their relationship with the environment. This survey provides a comprehensive review of the current state of human video generation, marking, to the best of our knowledge, the first extensive literature review in this domain. We start with an introduction to the fundamentals of human video generation and the evolution of generative models that have facilitated the field's growth. We then examine the main methods employed for three key sub-tasks within human video generation: text-driven, audio-driven, and pose-driven motion generation. These areas are explored concerning the conditions that guide the generation process.
Furthermore, we offer a collection of the most commonly utilized datasets and the evaluation metrics that are crucial in assessing the quality and realism of generated videos. The survey concludes with a discussion of the current challenges in the field and suggests possible directions for future research. The goal of this survey is to offer the research community a clear and holistic view of the advancements in human video generation, highlighting the milestones achieved and the challenges that lie ahead. 
\end{abstract}


\section{Introduction}
\IEEEPARstart{H}{\textbf{uman}} \textbf{video generation} task aims to synthesize natural and realistic \textbf{2D} human video sequences with generative models given control conditions such as text \cite{he2024id,ma2024follow}, audio \cite{qian2021speech,he2024co,zhu2021let,islam2019beat} and pose \cite{chang2023magicdance,animateanyone2024}. These generated video sequences feature full-body or half-body human figures, including detailed motion representations of body parts and faces. Recently, this field has gained significant attention due to a wide range of potential applications, including film production, video games, AR/VR, human-robot interaction, digital humans, and accessible human-machine interaction.

\begin{figure*}[h]
    \centering
    \includegraphics[width=0.8\textwidth]{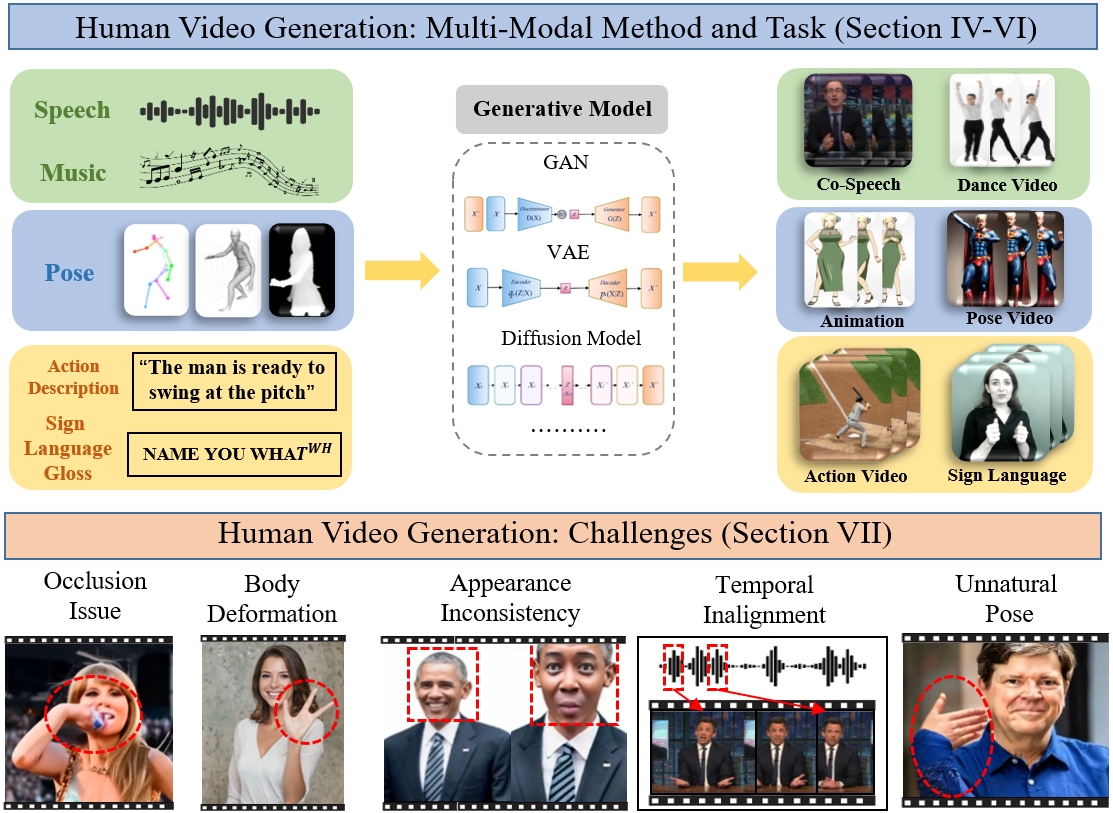}
        \caption{An overview of typical Multi-Condition human video generation methods and challenges.}
    \label{fig1:overview}
\end{figure*}

Recently, human video generation has achieved rapid progress benefiting from advancements in generation methods, \textit{i.e.,} 
Variational Autoencoders (VAE) \cite{kingma2013auto}, Generative Adversarial Networks (GAN) \cite{goodfellow2014generative}, and Diffusion Models \cite{ho2020denoising}.
However, studying such a video synthesis problem is known to be challenging for the following reasons. Firstly, the appearance consistency of humans along the time sequence is a significant obstacle in this task. Secondly, the deformation of the human body that people are sensitive to
in a synthesized video is hard to avoid, \textit{i.e.,} finger abnormalities, as shown in Fig. \ref{fig1:overview}.
 Thirdly, the complexity of human motion video extends beyond just modeling the face; it also involves accurately modeling body motion and maintaining background consistency and harmony with body parts. Additionally, the demand for human motion generation often includes a context as the condition, such as text description, audio signals, pose sequences, ensuring temporal alignment with these conditional signals is crucial for producing a coherent and realistic human video.

In response to the rapid development and emerging challenges of human video generation, we present a comprehensive survey of this field to help the community keep track of its progress. 

In summary, the main contributions of this survey are fourfold:
\begin{itemize}
    \item  We have carefully specified the boundaries of human video generation, offering a comprehensive analysis of recent advancements within this domain. We have categorized these advancements into three primary groups based on the modality driving the generation process: text-driven, audio-driven, and pose-driven. To our knowledge, this is the first survey that provides a systematic and focused examination of this particular field.

    \item We thoroughly examine the challenges and hurdles in human video generation through massive related methods and an extensive inventory of relevant datasets, challenges, evaluation metrics, and commercial projects. This paper guides readers in selecting suitable baselines or solutions for their unique applications. Additionally, our findings offer valuable insights into enhancing current methodologies.

    \item  Drawing from our detailed literature review and in-depth analysis, we have identified several promising directions for future development in human motion generation.

    \item  We also provide a continuously updated GitHub repository that includes the latest developments in the field, as well as links to awesome works and datasets. We aim to provide the research community the most cutting-edge information and provide easy access to important research works, datasets, and applications. For more details, please visit our \href{https://github.com/wentaoL86/Awesome-Human-Body-Video-Generation}{\textcolor{blue}{repository link}}
    
\end{itemize}

The survey is organized as follows. In Section \ref{sec:comparison}, we discuss the comparison with the previous survey. Section \ref{sec:data and metric} covers the fundamentals of the task, including the commonly used datasets and the evaluation metrics.
 In Sections \ref{sec:text}-\ref{section:pose}, we summarize existing approaches for human motion generation based on different conditional signals, respectively, including text, audio, and pose.
 Finally, we draw conclusions and provide insights for this field in Section \ref{sec:discussion}.

\section{Comparisons with Previous Surveys} 
\label{sec:comparison}
To the best of our knowledge, this survey is the first to focus directly on the human video generation task. Although several surveys have been conducted on video or motion generation, the differences between our survey and existing ones are mainly in the following three aspects. 

\textbf{1) Different Scope. }This survey focuses on human video generation, which is a 2D video generation task that uses a generative model to input text, audio, posture, or other modal data and uses full-body or half-body characters, including hands and faces as generated subjects. Compared with the general video generation task that many previous surveys~\cite{aldausari2022video,xing2023survey,cho2024sora,li2024survey} have focused on, this paper details the unique challenges and developments of human generation.
Additionally, surveys~\cite{chen2020comprises, chensurvey} concentrated solely on the talking head task, which focuses only on the generation of the head. However, the scope of this survey pays additional attention to hands, thus extending to the generation of half-body and full-body.
Furthermore, the work by Zhu ~\textit{et al.} ~\cite{zhu2023human} explicitly addresses motion generation, emphasizing human poses rather than video generation. 

\textbf{2) Video Perspective.} This paper especially discusses human generation challenges from a video perspective. In contrast, previous human generation surveys ~\cite{liao2024appearance,10.1145/3575656} focused on the problems in image generation. 

\textbf{3) New Insight.} To explore and solve the special challenges in human video generation and improve the generation quality, this paper provides a comprehensive analysis of the human video generation task through detailed methods and challenge discussion, as well as summarizes extra relevant datasets, evaluation metrics, and existing commercial projects. Our goal is to offer readers a clear and concise insight into the factors contributing to a successful human video generation and to answer the question, ``\textit{What Makes a Good Human Video Generation?}"

\begin{table*}[ht]
\centering
\setlength{\extrarowheight}{2pt}  
\begin{tabular}{>{\centering\arraybackslash}m{3.2cm}>{\centering\arraybackslash}m{5.5cm}>{\centering\arraybackslash}m{8cm}}
\hline
\textbf{Category} & \textbf{Metric} & \textbf{Description} \\ \hline
\multirow{8}{*}{\textbf{Image Quality}} 
& L1 Error & Measures the absolute pixel-level difference between predicted and ground truth frames. \\ \cline{2-3} 
& Peak Signal-to-Noise Ratio (PSNR) \cite{hore2010image} & Quantifies similarity between generated and real images in dB. \\ \cline{2-3} 
& Structural Similarity Index (SSIM) \cite{wang2004image} & Evaluate structural similarity considering luminance, contrast, and structure. \\ \cline{2-3} 
& Learned Perceptual Image Patch Similarity (LPIPS) \cite{zhang2018unreasonable}& A deep learning-based metric evaluating visual similarity. Lower LPIPS indicates higher similarity. \\ \cline{2-3} 
& Fréchet Inception Distance (FID) \cite{heusel2017gans} & Compares the feature distribution between generated and real samples. Lower FID indicates better quality. \\ \hline

\multirow{13}{*}{\textbf{Video Quality}} 
& Kernel Video Distance (KVD) \cite{unterthiner2018towards} & Measures the distribution distance between generated and real video sequences. \\ \cline{2-3} 
& Fréchet Video Distance (FVD) \cite{unterthiner2018towards} & Measures the distance between the distributions of generated and real videos. \\ \cline{2-3} 
& Average Content Distance (ACD) \cite{mocogan2018}& Assesses action sequence consistency in generated videos, especially for gesture generation tasks. \\ \cline{2-3} 
& Warping Error (WE) \cite{liu2024evalcrafter}& Obtain the optical flow of each two frames, then calculate the pixel-wise differences
between the warped image and the predicted image. \\ \cline{2-3} 
& Fréchet Gesture Distance (FGD) \cite{yoon2020speech} & Measure the distribution gap between real and generated gestures in the feature space. \\ \cline{2-3} 
& Fréchet Template Distance (FTD) \cite{yoon2020speech} & similar to the FGD , measuring the distribution similarity between the generated ones and
the real ones in the feature space \\ \cline{2-3}

& Fréchet Inception Distance for Videos (FID-VID) \cite{magicpose2023} & Measures the distribution distance between generated and real video frames, incorporating both spatial and temporal features. Lower FID-VID indicates better quality. \\ \hline

\multirow{7}{*}{\textbf{Consistency}} 
& Beat Consistency (BC) & Assesses temporal consistency in videos content with audio. \\ \cline{2-3} \
& CLIP-I score \cite{he2024id} & Measures the face structural similarity between the reference image and the generated video. \\ \cline{2-3} 
& Beat Alignment Score (BAS) \cite{li2021ai} & Evaluate the motion-music correlation in terms of the similarity between the kinematic beats and music beats.  \\ \cline{2-3} 
& Frame Consistency (FC) \cite{esser2023structure} & Assesses temporal consistency in videos by calculating cosine similarity between feature vectors of consecutive frames. \\ \hline

\multirow{3}{*}{\textbf{Diversity}} 
& Inception Score (IS) \cite{salimans2016improved} & Measures the diversity and clarity of generated images and sometimes used for video quality. \\ \cline{2-3} 
& Diversity (Div) \cite{liu2022learning} &  Calculates feature distance between generated gestures on average. \\ \hline

\multirow{1}{*}{\textbf{Aesthetics}} 
& Dover Score \cite{wu2023exploring}& Measures the overall quality of the generated video from both technical and
aesthetic perspectives \\ \hline

\multirow{4}{*}{\textbf{Pose Accuracy}} 
& Percentage of Correct Keypoints (PCK) \cite{yang2012articulated} & Measures the proportion of keypoints that are correctly localized within a specified threshold distance from the ground truth keypoints. \\ \cline{2-3}
& Average Keypoint Distance (AKD) \cite{siarohin2019first} & Evaluates the accuracy of human keypoints in generated videos by comparing distances to real keypoints. \\ \cline{2-3} 
& Missing Keypoint Rate (MKR) \cite{zhao2022thin} & Measures the proportion of missed keypoints during detection or generation. \\ \hline

\end{tabular}
\caption{Commonly Used Evaluation Metrics for Human Video Generation.}
\label{tab:metrics}
\end{table*}

\section{Dataset and Matrix}
\label{sec:data and metric}
\subsection{Metrics}
Comparing different methods in this field requires appropriate and comprehensive evaluation metrics. However, evaluating generated human videos presents significant challenges due to factors such as the one-to-many mapping nature, the subjectivity inherent in human evaluations, and the complexity of high-level conditional signals. To address these challenges, this section provides an overview of the most commonly used evaluation metrics, highlighting their advantages and limitations. The details of the metrics are shown in Table. \ref{tab:metrics}

We summarize that the evaluation of generated human videos in this field covers several critical aspects: Image Quality, Video Quality, Consistency, Diversity, Aesthetics, and Action Accuracy. Each of these categories is essential for a comprehensive assessment of the performance of different methods. 

\textbf{Image Quality} focuses on the visual fidelity of individual frames, evaluating pixel-level differences, structural similarity, and perceptual similarity to ensure frames closely match real ones. 

\textbf{Video Quality} extends this evaluation to the temporal domain, assessing the coherence and realism of frame sequences to capture the dynamic nature of real-world actions.

\textbf{Temporal Consistency} is to ensure that the generated content maintains a natural flow and synchronization over time, which is crucial for applications involving synchronized audio and video. 

\textbf{Diversity} is to evaluate the variety and richness of the generated content, ensuring the model can produce a wide range of realistic videos. 

\textbf{Action Accuracy} is to assess the precision of human actions and movements within the videos, which is vital for applications where the correctness of these actions is paramount. Together, these metrics provide a comprehensive framework for evaluating the performance and quality of methods in human video generation.

\begin{figure*}[h]
    \centering
    \includegraphics[width=0.8\textwidth]{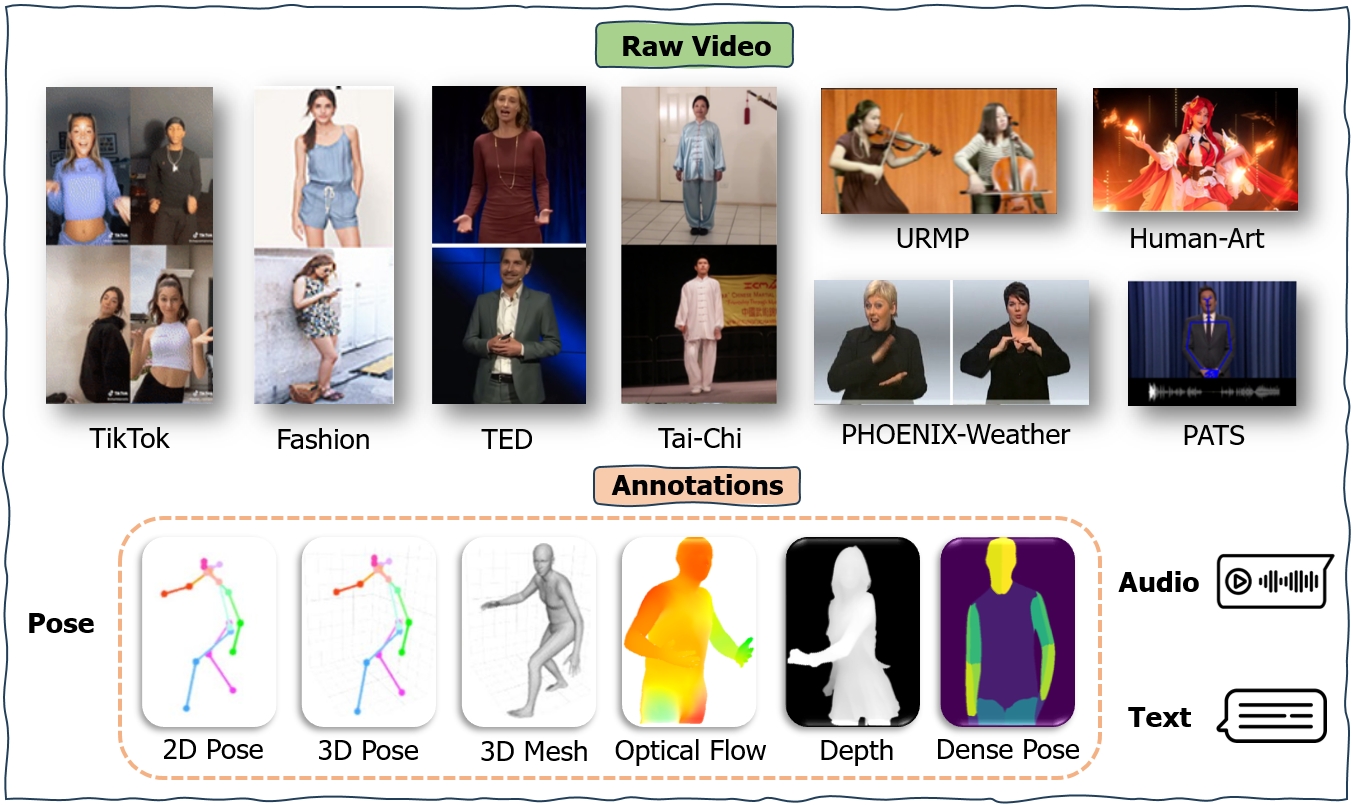}
        \caption{Some examples of human video datasets and annotation formats.}
    \label{fig1:dataset}
\end{figure*}

\begin{table*}[h!]
\centering
\begin{tabular}{>{\centering\arraybackslash}m{3cm}>{\centering\arraybackslash}m{3cm}>{\centering\arraybackslash}m{1.2cm}>{\centering\arraybackslash}m{2cm}>{\centering\arraybackslash}m{4.5cm}>{\centering\arraybackslash}m{1.2cm}}
\hline
\textbf{Category} & \textbf{Dataset Name} & \textbf{Year} & \textbf{Data Size} & \textbf{Modality} & \textbf{Source} \\ \hline

\multirow{8}{*}{\textbf{Human Action}} 
& ASTS \cite{gorelick2007actions} & 2005 & 90 videos& Video/Mask & \href{https://www.wisdom.weizmann.ac.il/~vision/SpaceTimeActions.html}{Link} \\ 
& UCF-101 \cite{soomro2012ucf101} & 2012 & $\sim$13k videos & Video & \href{https://www.crcv.ucf.edu/data/UCF101.php}{Link} \\ 
& human3.6m \cite{ionescu2013human3} & 2014 & 3.6M frames & Video/2D-Pose & \href{http://vision.imar.ro/human3.6m/description.php}{Link} \\ 
& NTU RGB+D \cite{shahroudy2016ntu}& 2016 & $\sim$114k videos& Video/3D-pose/Depth & \href{https://rose1.ntu.edu.sg/dataset/actionRecognition/}{Link} \\ 
& TaiChi \cite{siarohin2019first} & 2019 & 3k videos & Video & \href{https://github.com/AliaksandrSiarohin/first-order-model}{Link} \\ 

& HAR \cite{sharjeel2023har} & 2023 & $\sim$1k videos& Video & \href{https://www.kaggle.com/datasets/sharjeelmazhar/human-activity-recognition-video-dataset}{Link} \\ 
& 3D People Synthetic \cite{pumarola20193dpeople} & 2023 & $\sim$22k videos & Video/Masks/Pose/Depth/Mesh/OF & \href{https://drive.google.com/file/d/1N9gioWnkb3ZZytmT3Nzx4VjXjHxLsVB9/view}{Link} \\
& MSP-Avatar \cite{sadoughi2015msp} & 2023 & 74 videos & Video/Audio/Motion  & \href{https://ecs.utdallas.edu/research/researchlabs/msp-lab/MSP-AVATAR.html}{Link} \\
\hline

\multirow{6}{*}{\textbf{Human Dance }} 
& EverybodyDance \cite{chan2019everybody} & 2019 & 105 videos& Video/2D-Pose & \href{https://github.com/carolineec/EverybodyDanceNow}{Link} \\ 
& AIST++ \cite{li2021ai}  & 2021 & 10k videos & Video/Audio/3D-Pose & \href{https://google.github.io/aistplusplus_dataset/factsfigures.html}{Link} \\ 
& TikTok \cite{TikTok2021} & 2021 & 340 videos & Video/Depth/Mesh & \href{https://www.kaggle.com/datasets/yasaminjafarian/tiktokdataset}{Link} \\ 
& DanceIt \cite{guo2021danceit}  & 2021 & 154 videos & Video/Audio/2D-Pose & \href{https://github.com/iCVTEAM/DanceIt?tab=readme-ov-file}{Link} \\
& TikTok-v4 \cite{chang2023magicdance} & 2023 & 350 videos& Video/2D-Pose & \href{https://drive.google.com/file/d/1jEK0YJ5AfZZuFNqGGqOtUPFx--TIebT9/view}{Link} \\ 
& Disco \cite{disco2024} & 2024 & 700k frames & Video/2D-Pose/Mask & \href{https://drive.google.com/file/d/1N9gioWnkb3ZZytmT3Nzx4VjXjHxLsVB9/view}{Link} \\ 
 \hline

\multirow{2}{*}{\textbf{Music Performance}} 
& Sub-URMP \cite{chen2017deep} & 2017 & $\sim$81 frames& Video/Audio & \href{https://www.cs.rochester.edu/~cxu22/d/vagan/}{Link}\\
& URMP \cite{li2018creating} & 2018 & 44 videos& Video/Musical Score/Audio & \href{https://labsites.rochester.edu/air/projects/URMP.html}{Link} \\
 \hline

\multirow{3}{*}{\textbf{Human Fashion}} 
& DeepFashion \cite{liu2016deepfashion} & 2016 & $\sim$800k frames & Video/Mask/Text & \href{https://mmlab.ie.cuhk.edu.hk/projects/DeepFashion.html}{Link} \\            
& Fashion \cite{dwnet2019}  & 2019 & 600 videos& Video & \href{https://vision.cs.ubc.ca/datasets/fashion/}{Link} \\ 
& Fashion-Text2Video \cite{jiang2023text2performer} & 2023 & 600 videos& Video/Text & \href{https://drive.google.com/drive/folders/1NFd_irnw8kgNcu5KfWhRA8RZPdBK5p1I}{Link} \\ \hline

\multirow{1}{*}{\textbf{Human Art}} 
& HumanArt \cite{ju2023humanart} & 2023 & 50k frames & Video/Text/2D-Pose &  \href{https://idea-research.github.io/HumanArt/}{Link} \\ \hline

\multirow{8}{*}{\textbf{Body Language}} 
& MS-ASL \cite{msasl} & 2018 & ~25k videos & Video/Text & \href{https://www.microsoft.com/en-us/research/project/ms-asl/}{Link} \\
&PHOENIX14T\cite{camgoz2018neural} &2018&$\sim$68K frames&Video/Text  &  \href{https://www-i6.informatik.rwth-aachen.de/~koller/RWTH-PHOENIX-2014-T/}{Link} \\ 
&How2sign\cite{duarte2021how2sign} &2021&$\sim$35K frames&Video/Text/2D-Pose/Depth  &  \href{https://how2sign.github.io/}{Link} \\
& Bold \cite{luo2020arbee} & 2023 & $\sim$10k videos & Video//Text/Audio/3D-Pose &\href{https://cydar.ist.psu.edu/emotionchallenge/dataset.php}{Link} \\
 
& MCCS-2023\cite{liu2023cross,liu2024computation} &2023&$\sim$ 4k videos &Video/2D-Pose/3D-Pose/Text/Audio&\href{https://mccs-2023.github.io/}{Link}\\ 

& Speech2gesture \cite{ginosar2019learning} & 2019 & 60k& Video/Audio/2D-Pose & \href{https://github.com/amirbar/speech2gesture/blob/master/data/dataset.md}{Link} \\ 
& Pats \cite{liu2022audio} & 2020 & 84k videos & Videos/Text/Audio/2D-Pose & \href{https://cmu.app.box.com/s/obw6iazfrvoar11ymw01bxd7wxz2amzn}{Link} \\ 
& TED gesture \cite{yoonICRA19} & 2021 & $\sim$2k videos& Video/Text/2D-Pose/Audio & \href{https://github.com/youngwoo-yoon/youtube-gesture-dataset}{Link} \\ 
& Ted-talk \cite{siarohin2021motion}& 2021 & $\sim$3k videos& Video & \href{https://github.com/snap-research/articulated-animation}{Link} \\

\hline

\end{tabular}
\caption{Dataset Information for human video generation.}
\label{tab:datasets}
\end{table*}

\begin{figure*}[h]
    \centering
    \includegraphics[width=0.75\textwidth]{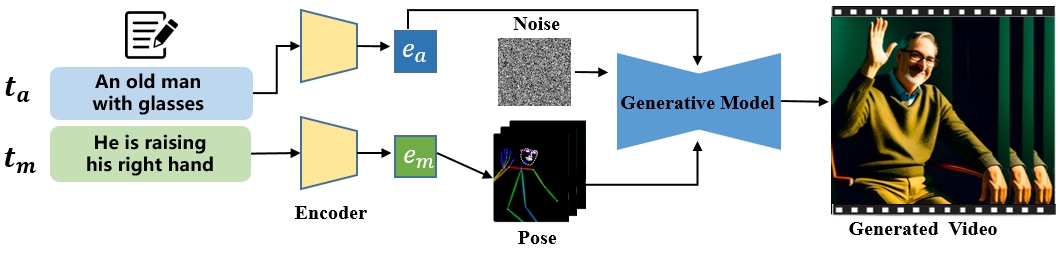}
        \caption{An overview of text to human video generation approaches.}.
    \label{fig:text-guided}
\end{figure*}

\subsection{Datasets}
Recently, various datasets have been utilized in the research on human video generation, encompassing a diverse array of scenes, actions, and backgrounds. The primary datasets include videos of dance, fashion, and daily activities, sourced from widely accessible platforms such as \textit{TikTok} and \textit{YouTube}. These datasets provide diverse data to support the training and evaluation of existing methods. The details of the datasets are shown in Table. \ref{tab:datasets}

For video generation tasks, effectively representing pose and motion information in videos is crucial. In this section, we will introduce common pose annotation formats, their characteristics, and commonly used methods.

\textbf{2D Pose} uses keypoints to form a skeletal graph for recognizing and analyzing human poses in 2D. Data is typically formatted as a set of \textit{(x, y)} coordinates for each joint. Common methods: OpenPose \cite{OpenPose2017}, DwPose \cite{yang2023effective}, PoseNet \cite{kendall2015posenet}, HRNet \cite{wang2020deep}, StackPose \cite{stackpose18}.

\textbf{3D Pose} adds depth to 2D poses, providing 3D coordinates \textit{(x, y, z)} for detailed human pose information. Common methods: ExPose \cite{choutas2020monocular}, Alphapose \cite{fang2022alphapose}, MotionBERT \cite{zhu2023motionbert}.

\textbf{3D Mesh} uses polygonal meshes to represent human surface shapes for realistic models. Data formats often include vertices and faces of the mesh. Common methods: SMPL\cite{loper2023smpl}, SMPL-X\cite{pavlakos2019expressive}.

\textbf{Optical Flow} represents motion vectors of pixels to describe motion direction and speed in videos. Data is typically stored as a 2D field of vectors. Common methods: MMFlow~\cite{mmflow2021}, FlowNet \cite{ilg2017flownet}, RAFT\cite{teed2020raft}.

\textbf{Depth} creates a depth map showing the distance of each pixel from the camera, useful for 3D reconstruction and AR. Data is usually in the form of depth images where each pixel value corresponds to the distance from the camera. Common methods: vid2depth \cite{mahjourian2018unsupervised}, monodepth2 \cite{godard2019digging}, Depth Anything~\cite{depthanything2024}.

\textbf{Dense Pose} maps 3D body surface coordinates to each pixel for detailed pose information. Data includes UV coordinates for each pixel mapped to a 3D body model. Common methods: DensePose \cite{densepose2018}.

\section{Text to Human Video Generation}
\label{sec:text}

In the following sections \ref{sec:text}-\ref{section:pose}, we will focus on the methods of human video generation based on different condition signals. Firstly, we will introduce the text-driven human video generation methods. 

Text can describe specific appearances, scenes, and styles, providing a rich source of information for generative models to control the generated content. Recent generative methods such as stable diffusion\cite{rombach2022high} and Sora \cite{cho2024sora} have shown that using text as input to generate images and videos has achieved impressive results.

However, different from the general video generation tasks which focus on the coherence of the video, human video generation requires precise control over both the appearance and movement of the human body.
Existing methods approach this challenge from two main angles: using text to maintain appearance and extracting semantic information from text to control poses. The overview of existing research in text-driven human video generation is shown in Fig. \ref{fig:text-guided}.

\subsection{Text-driven Human Appearance Control} 
To control the appearance of the human body in the generated video, there are two approaches: one is to directly provide \textbf{reference images}, and the other is to use input \textbf{text descriptions} to control the generated human appearance. Here, we discuss the text-driven human appearance control methods. To ensure the \textbf{consistency of appearance} in generated videos with the textual descriptions while preserving identity details during frames, ID-Animator \cite{he2024id} leverages a pre-trained text-to-video (T2V) model with a lightweight face adapter to encode identity-relevant embeddings. Text descriptions guide the generation of human videos and control the character's appearance in the video. Similarly, \cite{ma2024follow} uses text descriptions to provide semantic information about the content of the characters, ensuring the generated videos align with the textual descriptions. 

\subsection{Text-driven Human Motion Control} 
Existing methods for precisely controlling the motion of the human body in generated videos typically follow two approaches:

1) One approach follows a \textbf{two-stage pipeline}. It first generates corresponding poses based on the semantics of the input text according to the task and then uses these generated poses to guide the motion. More details about the pose-guided generation methods in the second stage can be referred in Section \ref{section:pose}. For this type of task, it is necessary to establish a connection between text and poses to control motion in a video.
HMTV \cite{kim2024human} uses descriptive text to generate initial 3D human motion and control camera angles, ensuring dynamic and realistic video outputs. The text guides the actions and camera movements in the video, providing precise control over the character's movements and the viewer's perspective. For the Sign Language Production task, SignSynth \cite{stoll2020signsynth} uses a Gloss2Pose network to generate sign language poses and a GAN to create high-quality sign language videos. Similarly, H-DNA \cite{natarajan2022development} translates spoken sentences into sign language videos by first generating sign gesture poses and then using a GAN to produce the final video. In SignLLM \cite{fang2024signllm}, text descriptions are converted into gloss (an intermediate sign language representation) and then mapped to poses, which are rendered into Sign Language videos. Here, the semantics of the text are captured to align with the described human pose. In Cued Speech \cite{cornett1967cued,liu2019pilot} Generation task, \cite{lei2024bridge} first leveraged a Large Language Model (LLM) to convert text into a descriptive gloss and then used the gloss to generate a fine-grained pose.  


2) The other approach directly uses text as a prompt to guide the generation of video actions. For instance, Text2Performer \cite{jiang2023text2performer} involves the motion text and a motion encoder. motion text describes the movement, such as \textit{``She is swinging to the right."} The model implicitly models these descriptions by separately representing appearance and motion, thereby generating high-quality videos with consistent appearance and actions.

\begin{figure*}[h]
    \centering
    \includegraphics[width=0.8\textwidth]{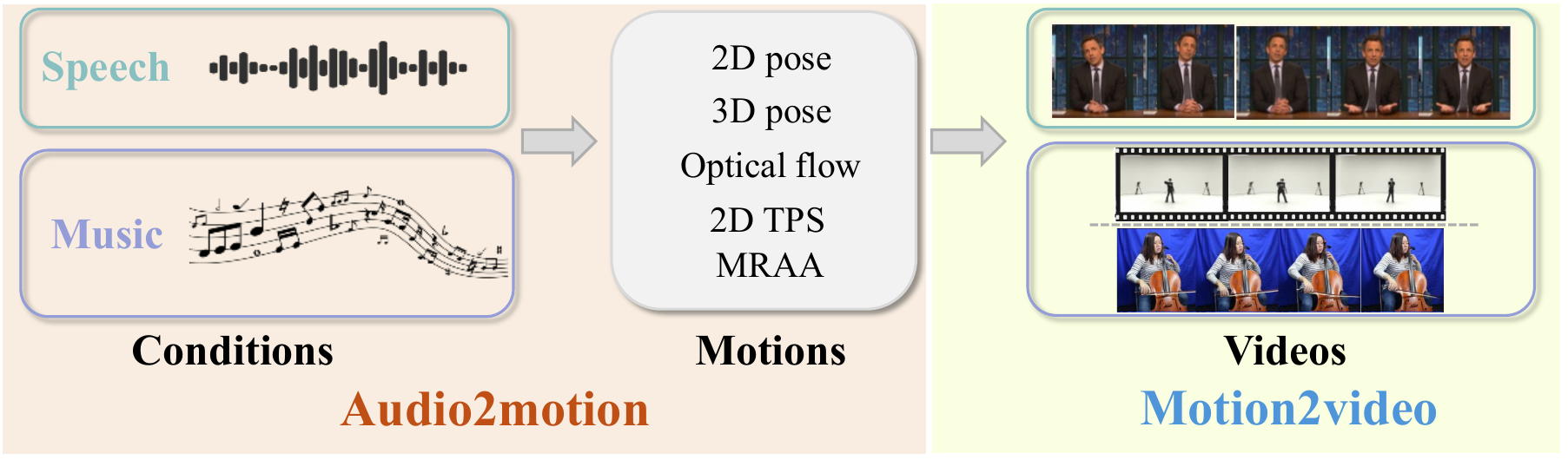}
        \caption{An overview of audio to human video generation approaches. Example images adapted from \cite{he2024co,wang2024dance,jia2023music2play}.}.
    \label{fig:a2v_overview}
\end{figure*}

\section{Audio to Human Video Generation}
\label{sec:audio}

In addition to textual descriptions, human video generation from audio signals has also been explored in this survey. In this section, we mainly discuss two main subtasks: \textbf{speech-driven human video} and \textbf{music-driven human video}. 
Speech-driven human video generation aims to generate a sequence of human gestures based on input speech audio, which requires the generated human motion to be harmonious with the audio, not only in terms of high-level semantics but also emotion and rhythm. 
While music-driven human video generation focuses on synthesizing the video of a person dancing or playing a certain instrument guided by a given music clip, which especially lies in the low-level beat alignment.
In this scenario, the direct conversion of audio into video poses a complex challenge. Previous research has often followed a two-stage pipeline, including audio-to-motion and motion-to-video, as illustrated in Fig. \ref{fig:a2v_overview}.

\begin{table*}
\centering
\begin{tabular}{cccccc}
\hline
\textbf{Condition} &\textbf{Method} & \textbf{Venue} & \textbf{Model}   &\textbf{Motion Feature}&\textbf{Dataset}\\
\hline
\multirow{7}{*}{Speech}& 
speech2gesture \cite{ginosar2019learning} & CVPR 2019& GAN& 2D pose&Speech2Gesture \cite{ginosar2019learning}\\

&Speech2video \cite{liao2020speech2video}& ACCV 2020& LSTM& 3D pose&Self-collection\\

&Qian \textit{al.} \cite{qian2021speech}& ICCV 2021& VAE& 2D pose&Speech2Gesture \cite{ginosar2019learning}\\

&ANGIE \cite{liu2022audio}&NeurIPS 2022&VQ-VAE& MRAA&PATS \cite{liu2022audio}\\

&DR$^2$ \cite{zhang2024dr2}&WACV 2024&VAE&3D pose&Speech2Gesture \cite{ginosar2019learning}, Self-collection\\

&DiffTED \cite{hogue2024diffted}&CVPR 2024&Diffusion& 2D TPS &TED-talks \cite{siarohin2021motion} \\

 &He \textit{al.} \cite{he2024co}& CVPR 2024& Diffusion& 2D TPS, optical flow&PATS \cite{liu2022audio} \\
 \hline
\multirow{5}{*}{Music}& 
 Islam \textit{al.} \cite{islam2019beat}&RAAICON 2019&GAN&2D pose&Self-collection\\

&DanceIt \cite{guo2021danceit}&TIP 2021&GAN& 2D pose &Self-collection\\

&Dabfusion \cite{wang2024dance}& ArXiv 2024&Diffusion&Optical flow &AIST++ \cite{li2021ai}\\

 &Zhu \textit{al.} \cite{zhu2021let}& ICPR 2021&CNN& keypoint&Sub-URMP \cite{chen2017deep}\\

  &Music2Play \cite{jia2023music2play}&CAC 2023& LSTM& 2D pose, optical flow &URMP \cite{li2018creating}\\
   \hline

\end{tabular}
\caption{Summary of works related to audio to human video generation. }
    
\end{table*}

\subsection{Speech-driven Human Video Generation}
Many existing works have concentrated on generating talking videos, primarily focusing on the head region \cite{gao2023high,gan2023efficient}. In contrast, our review focuses on works that include body gestures \cite{ginosar2019learning,liao2020speech2video,qian2021speech,zhang2024dr2,hogue2024diffted,he2024co}. To the best of our knowledge, all of these works fall under the field of \textbf{co-speech gesture} video generation. Given the importance of motion representation for the final video, we review these works from the perspective of motion generation.

In speech-driven human video generation, some methods \cite{ginosar2019learning,liao2020speech2video,zhang2024dr2} synthesize talking videos from sequences of 2D skeletons \cite{ginosar2019learning,qian2021speech} or 3D models \cite{liao2020speech2video,zhang2024dr2}, with the rendering process being separate from the generation of the gestures. However, hand-crafted structural human priors like 2D/3D skeletons completely discard appearance information around key points, making precise motion control and video rendering highly challenging. Additionally, the pre-training of pose estimators relies on hand-crafted annotations, leading to error accumulation and often resulting in jitters.
To alleviate these issues, ANGIE \cite{liu2022audio} utilizes an unsupervised feature, MRAA \cite{siarohin2021motion}, to model body motion. A VQ-VAE \cite{van2017neural} is then used to quantize common patterns, followed by a GPT-like network that predicts discrete motion patterns to generate gesture videos. However, MRAA, being a coarse modeling of motion, is linear and fails to represent complex-shaped regions, limiting the quality of gesture videos generated by ANGIE. Additionally, directly associating covariance with speech is inappropriate.
To address these challenges, DiffTED and He al. propose decoupling motion from gesture videos while preserving critical appearance information of body regions. They use the learned 2D keypoints of the Thin-plate Spline (TPS) motion model \cite{zhao2022thin} as targets for generation and leverage the TPS motion model to render the keypoints into images. Additionally, motivated by the success of recent diffusion models \cite{ho2020denoising}, DiffTED and He \textit{al.} propose a diffusion-based approach to generate diverse gesture sequences.

\subsection{Music-driven Human Video Generation}
 Music-driven human video generation uniquely intersects motion synthesis and music interpretation, aiming to create human motions synchronized with input music beat. This extends beyond general motion synthesis, as beat-aligned motions are complex to animate \cite{wang2024dance}.
 We have explored two sub-tasks, \textit{i.e.,} \textbf{music-to-dance} and \textbf{music-to-performance}.
 
To achieve beat sensing motion generation, 
some music-to-dance video generation works \cite{islam2019beat,wang2024dance} explicitly detect beat from music audio, or design a matching phase learns the relationship between these two different modalities \cite{guo2021danceit}.
Islam \textit{al.} \cite{islam2019beat} perform beat detection and repeated pattern extraction from input music first and then generate mathematical models of a person dancing and convert them into realistic images of the target person.
Dabfusion \cite{wang2024dance} applies a beat extractor to explicitly disentangle beat features from music. These beat features are then used to guide the production of latent optical flows, followed by backward flow estimation to generate the output video.
Differently, DanceIt \cite{guo2021danceit} learns the relationship between these two different modalities at the first matching phase, then retrieves a sequence of pose fragments for each music audio and performs spatial-temporal alignment at the generation phase.

For music-to-performance video generation, it is challenging to generate high-dimensional temporal consistent videos from low-dimensional audio modality. Zhu \textit{al.} \cite{zhu2021let} propose a multi-staged framework that first generates the coarse video from given audio and then makes refinements by integrating intra-frame structure information from predicted keypoints and temporal information for final performance video generation. Music2Play \cite{jia2023music2play} gains a sequence of poses in an auto-regressive way and, estimates the dense flow field information from the pair of poses, finally fuses multi-modal information (audio, flow, and image) to synthesize the output frame.

\section{Pose to Human Video Generation}
\label{section:pose}

\begin{figure}[h]
    \centering
    \includegraphics[width=0.5\textwidth]{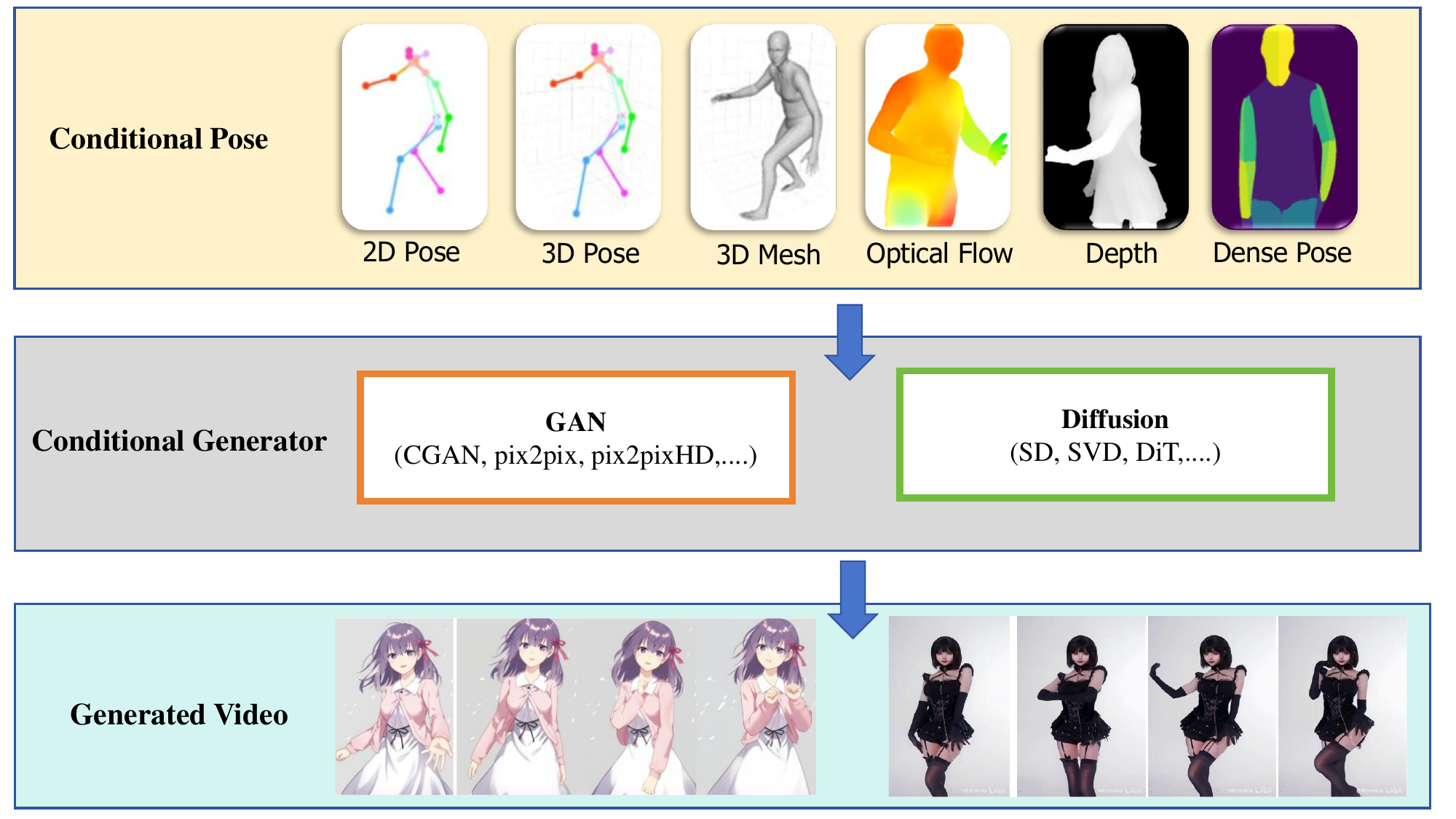}
        \caption{An overview of pose-guided human video generation approaches. Examples come from~\cite{vividpose2024} and~\cite{animateanyone2024}.}.
    \label{fig:pose}
\end{figure}

As illustrated in Fig.~\ref{fig:pose}, existing research in pose-driven human video generation has often followed a common pipeline. In the task of pose-driven human video generation, various pose types, including \textbf{skeleton pose, dense pose, depth, mesh, and optical flow} (as shown in Tab.~\ref{tab:pose}), serve as common guiding modalities along with the more traditional text and speech inputs.  According to the number of conditional poses, we can divide the existing pose guided human video generation methods into two categories. The first category uses only a single type of pose, which is recorded as \textbf{single-condition pose-guided methods}. The second category uses different types of pose signals, which are referred to as \textbf{multi-condition poses-guided methods}.

\subsection{Single-condition Pose-guided Methods}
Among all types of conditional signals, the most common are skeleton pose and dense pose.   Early pose-guided human video generation methods~\cite{mocogan2018,posevideo2018,poseanimation2021,HumanVideo2018,disentangled2019,dwnet2019,everydance2019,longhuman2020,Twostreamvan2020,mmflow2021,SGWGAN2021} based on GANs primarily utilized conditional adversarial networks such as CGAN~\cite{cgan}, pix2pix~\cite{pix2pix}, and pix2pixHD~\cite{pix2pixHD}.   These methods extracted skeleton poses using OpenPose~\cite{OpenPose2017} or StackPose~\cite{stackpose18} methods, or extracted dense pose using the DensePose method, and used the extracted skeleton pose or dense pose as conditional signal into CGAN or pix2pix generation models.

With the development of conditional generation models, current methods~\cite{magicpose2023,motionfollower2024,mimicmotion2024,animateanyone2024,unianimate2024} mostly utilize stable diffusion (SD)~\cite{LDMs} or Stable Video Diffusion (SVD)~\cite{SVD,alignlatents} as the backbone for video generation models.      For instance, the MagicPose~\cite{magicpose2023} injects pose features into the diffusion model by ControlNet~\cite{ControlNet}.  In contrast to directly utilizing ControlNet, methods such as MotionFollower~\cite{motionfollower2024}, MimicMotion~\cite{mimicmotion2024}, AnimateAnyone~\cite{animateanyone2024}, and UniAnimate~\cite{unianimate2024} extract skeleton poses from targvideo frames using DwPose~\cite{DWPose2023} or OpenPose~\cite{OpenPose2017}.      To align the extracted skeleton poses with the noise in the latent space and effectively leverage pose guidance during denoising processing, they design lightweight neural networks (composed of only a few convolutional layers) as pose guider.

Unlike the above skeleton pose-guided video generation diffusion models, methods like DreamPose~\cite{dreampose2023} and MagicAnimate~\cite{magicanimate2024} utilize the DensePose~\cite{densepose2018} method to extract dense pose and directly concatenate dense pose and noise into the denoising UNet by ControlNet.  Different from these types of 2D poses (skeleton pose and dense pose), Human4DiT~\cite{human4dit2024} extracts corresponding 3D mesh maps using SMPL~\cite{SMPL2023}.  Inspired by the work of Sora and other variants~\cite{cho2024sora,Latte2024}, Human4DiT~\cite{human4dit2024} regards Diffusion Transformer as the backbone for video generation.

\begin{table*}
\centering
\begin{tabular}
{>
{\centering\arraybackslash}m{0.6cm}>{\centering\arraybackslash}m{3cm}>
{\centering\arraybackslash}m{1.7cm}>{\centering\arraybackslash}m{2.5cm}>{\centering\arraybackslash}m{3.5cm}>{\centering\arraybackslash}m{3.5cm}}
\hline
Model                       & Method              & Venue    & Condition                            & Extractor                      &  Dataset                                                   \\ \hline
\multirow{10}{*}{GAN} 
& Cai~\textit{et al.} ~\cite{HumanVideo2018}      & ECCV 2018 & SK                      & SP~\cite{stackpose18}    & Human3.6M~\cite{ionescu2013human3}                                                  \\
& Yang~\textit{et al.}~\cite{posevideo2018}         & ECCV 2018 & SK                            & OP~\cite{OpenPose2017} &~\cite{soomro2012ucf101,actions2007,ck2010}
                            \\
                            & Yang~\textit{et al.}~\cite{disentangling2024}                & ICMEW 2019 &  SK            & OP~\cite{OpenPose2017}        & self-collection                                                             \\
                            & Chan ~\textit{et al.}~\cite{everydance2019}           & ICCV 2019 & SK                        & OP~\cite{OpenPose2017}                       & EverybodyDance \cite{chan2019everybody}                                                             \\
                            & DwNet~\cite{dwnet2019}                & BMVC 2019    & DS  & DP~\cite{densepose2018}                           & Fashion \cite{dwnet2019}                                               \\
                            & Naoya~\textit{et al.}~\cite{longhuman2020}               & ECCVW 2020    & SK          & OP~\cite{OpenPose2017} & Human3.6M~\cite{ionescu2013human3}                                                 \\
                            & Yoon~\textit{et al.}~\cite{poseanimation2021}          & CVPR 2021    &  DS               & DP~\cite{densepose2018}         & 3D-people~\cite{pumarola20193dpeople} \\
                            & SGW-GAN~\cite{SGWGAN2021}          & ArXiv  2021  & SK                        & OP~\cite{OpenPose2017}                       & MS-ASL~\cite{msasl}                                 \\
                           \midrule
\specialrule{0em}{1.5pt}{1.5pt}
\midrule
\multirow{16}{*}{Diffusion} & DreamPose~\cite{dreampose2023}            & ICCV 2023 & DS                          & DP~\cite{densepose2018}                       & Fashion~\cite{dwnet2019}                                                         \\
                            & LEO~\cite{leo2023}                  & ArXiv 2023   & OF                           & LIA~\cite{LIA2022}                            &           
                            \cite{TaichiHD2019,FaceForensics2018,CelebV-HQ2022}
                            \\
                            & DreaMoving~\cite{dreamoving2023}          & ArXiv 2023    & SK, DP                & DwP~\cite{DWPose2023}, ZoeDepth~\cite{ZoeDepth2023}               & self-collection                           \\
                            & DisCo~\cite{disco2024}               & CVPR 2024 & BG, SK           & G-SAM~\cite{GroundedSAM2023}, OP~\cite{OpenPose2017}        & TikTok~\cite{TikTok2021}                                                             \\
                            & Animate Anyone~\cite{animateanyone2024}      & CVPR 2024 & SK                        & DP~\cite{densepose2018},OP~\cite{OpenPose2017}             & self-collection                                                  \\
                            & MagicPose~\cite{magicpose2023}            & ICML 2024 & SK                       & OP~\cite{OpenPose2017}                       & TikTok~\cite{TikTok2021}                                                             \\
                            & MagicAnimate~\cite{magicanimate2024}        & CVPR 2024 & DS                           & DP~\cite{densepose2018}                      &    TikTok~\cite{TikTok2021}, TED-talks~\cite{siarohin2021motion}                           \\
                            & Champ~\cite{champ2024}              & ArXiv 2024   & DP, NM, SM, SK  & SMPL~\cite{SMPL2023}                           & self-collection                                                   \\
                            & PoseAnimate~\cite{poseanimate2024}          & ArXiv 2024   & SK                       & OP~\cite{OpenPose2017}                       & Training-Free                                                      \\
                            & Liu \textit{et al.}~\cite{disentangling2024}               & ArXiv 2024   & SK, BG           & DwP~\cite{DWPose2023}], HM~\cite{humanmatting2020}   & self-collection                                                 \\
                            & MotionFollower       & ArXiv 2024   & SK                        & DwP~\cite{DWPose2023}                         & self-collection                                                  \\
                            & Human4DiT~\cite{human4dit2024}           & ArXiv 2024   & MS                                & SMPL~\cite{SMPL2023}                           & self-collection
                            \\
                            & VividPose~\cite{vividpose2024}        & ArXiv  2024  & SK,  MS        & DwP~\cite{DWPose2023}, SMPL-X~\cite{smpler-x2024}               & TikTok~\cite{TikTok2021}                                                             \\
                            & UniAnimate~\cite{unianimate2024}        & ArXiv 2024   & SK                       & DwP~\cite{DWPose2023}                         & TikTok~\cite{TikTok2021}, Fashion~\cite{dwnet2019}                                                   \\
                            & FYP v2~\cite{follow-your-posev2}
                            & ArXiv 2024    & SK, DP, OF  & DwP~\cite{DWPose2023}, DA~\cite{depthanything2024}, MF~\cite{mmflow2021} & self-collection              \\
                            & MimicMotion~\cite{mimicmotion2024}         & ArXiv 2024   & SK                       & DwP~\cite{DWPose2023}                         & self-collection                                                \\ \hline
\end{tabular}
\caption{ List of Methods Focusing on Pose Guided Human Video Generation. SP, DP, OP and DwP represent StackPose~\cite{stackpose18}, DensePose~\cite{densepose2018}, OpenPose~\cite{OpenPose2017} and DwPose~\cite{DWPose2023}. G-SAM, DA, HM and MF represent Grounded-SAM~\cite{GroundedSAM2023}, Depth Anything~\cite{depthanything2024}, Human Matting~\cite{humanmatting2020} and MMFlow~\cite{mmflow2021}. SK, DS, MS, and OF represent skeleton pose, dense pose, mesh, and optical flow. BG, NM, and SM represent the background, normal map, and semantic map.}
\label{tab:pose}
\end{table*}

\subsection{Multi-condition Poses-guided Methods}
In addition to the single conditional pose-based human video generation, the recent success of SD~\cite{LDMs} and SVD~\cite{SVD,alignlatents} has laid the foundation for multi-conditional pose-guided human video generation. 
Most existing pose-guided methods use either skeleton pose or dense pose as the conditional input.  However, these single-condition pose-guided methods often exhibit poor generalization to complex backgrounds and suffer from occlusion issues between different bodies and parts of the same individual. 

To address the poor generalization considering the \textbf{complex backgrounds}, DISCO~\cite{disco2024} presents an innovative model architecture featuring disentangled control over background and skeleton pose, thereby improving the compositionality of dance generation. This architecture enables the integration of both seen and novel subjects, backgrounds, and poses from diverse sources. Follow-Your-Pose v2~\cite{follow-your-posev2} integrates an optical flow guide with other condition guiders to enhance background stability. Liu ~\textit{et al.} ~\cite{disentangling2024} separates the motion representations of the foreground and background, animating human figures with pose-based motion while modeling background motion using sparse tracking points to capture natural interactions between the figure's activity and environmental changes.

To tackle the \textbf{occlusion issues}, Follow-Your-Pose v2~\cite{follow-your-posev2} addresses occlusions in multi-character animation with a depth guider, and improves character appearance learning with a reference pose guider. VividPose~\cite{vividpose2024} introduces depth and mesh information, particularly in conjunction with the SMPL-X~\cite{smpler-x2024} model, which helps the system to better handle occlusions and complex movements that are common in human pose sequences. DreaMoving~\cite{dreamoving2023} integrates depth information and skeleton pose,  helping the model to understand the spatial relationships between different parts of the body and the environment.
The depth information is useful for handling occlusions as it allows the model to determine which body parts are in front of or behind others.

\section{Challenges}
In this section, we summarize the key challenges in the human video generation task, discuss the special challenges existing in the models guided by the particular modality, and explain the common problems faced by this task and related video generation tasks. Representative challenges include:  

\textbf{1) Occlusion Issue.}  In the collected videos, overlapping body parts or multiple people occlusion is common, but
most models cannot handle the problem of mutual influence well~\cite{follow-your-posev2,vividpose2024}.

\textbf{2) Body Deformation.} Ensuring that generated video features such as body shape, face, and hands adhere to typical human characteristics is a significant obstacle in this task. One common example of this issue is the occurrence of malformed hands~\cite{lu2023handrefinerrefiningmalformedhands}.


\textbf{3) Appearance Inconsistency.} The generation of human videos also requires that the various features of the human appearance, including face, body, clothing, accessories, etc., be consistent in the generated videos. However, most models cannot achieve utterly satisfactory consistency. 

\textbf{4) Background Influence.} When generating videos with the human body in the foreground, the consistency of the background and the harmony with the foreground human body is also a major challenge. Poor background control will affect the quality of human generation and bring additional jitter and distortion. 

\textbf{5) Temporal Inalignment.} In models guided by temporal signals, especially the audio-to-human video generation models, the synchronization of lips and voice is a significant challenge to improving the quality.


\textbf{6) Unnatural Pose.} Current generated human video often suffers from the unnatural pose problem. The specific manifestations of this problem include the inconsistency between the generated video and the inputted guided pose, as well as the naturalness of the movements in the generated videos. 

In addition to the representative challenges mentioned above, in text- or audio-driven models, due to the one-to-many mapping nature in the dataset, meaning that a single input text or audio can correspond to several valid outputs. As a result, attempting to directly match the input with a single 'correct' gesture can lead to an unreliable and biased association. This approach hinders the model's ability to capture and learn the variations present within the data~\cite{qian2021speech}.

It should be noted that since human video generation is essentially a branch of video generation, the efficiency challenges brought by the common use of diffusion models, the challenges of multi-view generation, and the challenges of high-resolution generation still have a significant impact on the generation quality.

\section{Conclusion and Discussion}
\label{sec:discussion}

\subsection{Conclusion}
In this survey, we provide a comprehensive overview of recent advancements in human video generation. Despite the rapid progress in this field, significant challenges remain that warrant further exploration. We summarize available dataset resources and commonly used evaluation metrics. Subsequently, we classify the existing researches based on conditional signals (\textit{i.e.} text, audio and pose) and discuss each category in detail.

\subsection{Discussion}

In this section, we aim to discuss in detail the factors influencing the quality of human video generation, excluding dataset scale.    To this end, we will focus on three aspects: generation paradigm, backbone, and condition pose.

\begin{itemize}
    \item \textbf{Generation Paradigm.} Compared to pose-driven methods (which can be regarded as one-stage methods), text and audio-driven methods can be divided into one-stage and two-stage approaches.    The former directly uses input text or audio as prompts to guide human video generation, while the latter generates poses from the input text or audio and then uses these generated poses as signals to guide human video generation. The introduction of various pose types, such as skeleton poses, in two-stage methods, provides additional geometric and semantic information, enhancing the accuracy and realism of video motions. This makes two-stage methods significantly more effective than one-stage methods, albeit at the cost of some efficiency.

\item \textbf{Backbone.} Diffusion models, such as SD and SVD, are widely used in various generative tasks, including human video generation, due to their superior performance and diversity. However, unlike GANs, which generate samples in a single sampling step, diffusion models require multiple sampling steps, thereby increasing the time cost for training and inference.

\item \textbf{Condition Pose.} Different types of conditional poses work because they provide complementary information.    The most common skeleton pose accurately describes the spatial information of the human body in the frame and the relative positions of body parts.    However, it captures discrete pose changes rather than continuous motion details, providing limited temporal coherence.    In contrast, optical flow inherently includes temporal information, capturing changes between consecutive frames and providing continuous motion trajectories in the feature space.    This allows the model to generate videos with smooth transitions between frames, avoiding jumps or discontinuities.    Moreover, the skeleton pose does not include background and detail modeling, whereas depth maps capture distance information between human body and the background, along with surface details and depth changes.    3D meshes offer detailed geometric structures of object surfaces that skeleton poses lack. In summary, different types of poses provide complementary spatiotemporal information, and there is no unified pose type that fulfills all requirements. Different scenarios and problems may require different poses.
\end{itemize}




\subsection{Future Work}
We outline several promising future directions from various perspectives, aiming to inspire new breakthroughs in human video generation research.

\begin{itemize}
    \item \textbf{Large-Scale High-Quality Human Video Datasets.} 
Existing public datasets, including those in the fields of human action and human dance, are relatively small in scale. Collecting high-quality human video datasets is both challenging and expensive. However, a large-scale, high-quality human video dataset is crucial for developing a foundational model for human video generation.
    \item \textbf{Long Video Generation.} 
Current human video generation methods typically produce videos lasting only several seconds. Generating videos that extend to several minutes or even hours presents a significant challenge. Therefore, future research should focus on the generation of long-duration human videos.
\item \textbf{Photorealistic Video Generation.} As previously mentioned, challenges such as occlusion, body deformation, pose unnaturalness, and appearance inconsistency can result in low-quality video generation. Resolving these visual and aesthetic issues to ensure that the generated human body movements follow real-world physical laws is a major challenge. Creating videos with highly realistic visual effects remains a difficult task.
\item \textbf{Human Video Diffusion Efficiency.} Diffusion models have become the backbone for human video generation tasks. However, the heavy training costs and deployment requirements of video diffusion models pose significant challenges. Reducing training costs and scaling down model size are crucial issues. Therefore, exploring the efficiency of video diffusion models is a valuable direction for future research.
    \item \textbf{Fine-Grained Controllability.} Existing multimodal-driven human video generation methods, even when incorporating additional, conditional signals such as 3D mesh and depth map alongside skeleton pose, still lack fine-grained control over specific body parts, particularly hands, and face. Future research could focus on achieving fine-grained, controllable generation of these detailed human body regions.
    \item \textbf{Interactivity.}
    In addition to exploring fine-grained controllability, future work could further investigate interactive controllability. This would allow users to manipulate elements such as arm movements or facial expressions through simple actions like clicking, ultimately generating human videos that meet user satisfaction.
    
\end{itemize}

\bibliographystyle{IEEEtran}
\bibliography{references}

\begin{thebibliography}{100}
\providecommand{\url}[1]{#1}
\csname url@samestyle\endcsname
\providecommand{\newblock}{\relax}
\providecommand{\bibinfo}[2]{#2}
\providecommand{\BIBentrySTDinterwordspacing}{\spaceskip=0pt\relax}
\providecommand{\BIBentryALTinterwordstretchfactor}{4}
\providecommand{\BIBentryALTinterwordspacing}{\spaceskip=\fontdimen2\font plus
\BIBentryALTinterwordstretchfactor\fontdimen3\font minus \fontdimen4\font\relax}
\providecommand{\BIBforeignlanguage}[2]{{%
\expandafter\ifx\csname l@#1\endcsname\relax
\typeout{** WARNING: IEEEtran.bst: No hyphenation pattern has been}%
\typeout{** loaded for the language `#1'. Using the pattern for}%
\typeout{** the default language instead.}%
\else
\language=\csname l@#1\endcsname
\fi
#2}}
\providecommand{\BIBdecl}{\relax}
\BIBdecl

\bibitem{he2024id}
X.~He, Q.~Liu, S.~Qian, X.~Wang, T.~Hu, K.~Cao, K.~Yan, M.~Zhou, and J.~Zhang, ``Id-animator: Zero-shot identity-preserving human video generation,'' \emph{arXiv}, 2024.

\bibitem{ma2024follow}
Y.~Ma, Y.~He, X.~Cun, X.~Wang, S.~Chen, X.~Li, and Q.~Chen, ``Follow your pose: Pose-guided text-to-video generation using pose-free videos,'' in \emph{AAAI}, 2024.

\bibitem{qian2021speech}
S.~Qian, Z.~Tu, Y.~Zhi, W.~Liu, and S.~Gao, ``Speech drives templates: Co-speech gesture synthesis with learned templates,'' in \emph{ICCV}, 2021.

\bibitem{he2024co}
X.~He, Q.~Huang, Z.~Zhang, Z.~Lin, Z.~Wu, S.~Yang, M.~Li, Z.~Chen, S.~Xu, and X.~Wu, ``Co-speech gesture video generation via motion-decoupled diffusion model,'' in \emph{CVPR}, 2024.

\bibitem{zhu2021let}
H.~Zhu, Y.~Li, F.~Zhu, A.~Zheng, and R.~He, ``Let's play music: Audio-driven performance video generation,'' in \emph{ICPR}, 2021.

\bibitem{islam2019beat}
M.~S. Islam, M.~S. Rahman, and M.~A. Amin, ``Beat based realistic dance video generation using deep learning,'' in \emph{RAAICON}, 2019.

\bibitem{chang2023magicdance}
D.~Chang, Y.~Shi, Q.~Gao, J.~Fu, H.~Xu, G.~Song, Q.~Yan, X.~Yang, and M.~Soleymani, ``Magicdance: Realistic human dance video generation with motions \& facial expressions transfer,'' \emph{arXiv}, 2023.

\bibitem{animateanyone2024}
L.~Hu, ``Animate anyone: Consistent and controllable image-to-video synthesis for character animation,'' in \emph{CVPR}, 2024.

\bibitem{kingma2013auto}
D.~P. Kingma and M.~Welling, ``Auto-encoding variational bayes,'' \emph{arXiv preprint arXiv:1312.6114}, 2013.

\bibitem{goodfellow2014generative}
I.~Goodfellow, J.~Pouget-Abadie, M.~Mirza, B.~Xu, D.~Warde-Farley, S.~Ozair, A.~Courville, and Y.~Bengio, ``Generative adversarial nets,'' \emph{NeurIPS}, 2014.

\bibitem{ho2020denoising}
J.~Ho, A.~Jain, and P.~Abbeel, ``Denoising diffusion probabilistic models,'' \emph{NeurIPS}, 2020.

\bibitem{aldausari2022video}
N.~Aldausari, A.~Sowmya, N.~Marcus, and G.~Mohammadi, ``Video generative adversarial networks: a review,'' \emph{CSUR}, vol.~55, no.~2, pp. 1--25, 2022.

\bibitem{xing2023survey}
Z.~Xing, Q.~Feng, H.~Chen, Q.~Dai, H.~Hu, H.~Xu, Z.~Wu, and Y.-G. Jiang, ``A survey on video diffusion models,'' \emph{arXiv}, 2023.

\bibitem{cho2024sora}
J.~Cho, F.~D. Puspitasari, S.~Zheng, J.~Zheng, L.-H. Lee, T.-H. Kim, C.~S. Hong, and C.~Zhang, ``Sora as an agi world model? a complete survey on text-to-video generation,'' \emph{arXiv}, 2024.

\bibitem{li2024survey}
C.~Li, D.~Huang, Z.~Lu, Y.~Xiao, Q.~Pei, and L.~Bai, ``A survey on long video generation: Challenges, methods, and prospects,'' \emph{arXiv}, 2024.

\bibitem{chen2020comprises}
L.~Chen, G.~Cui, Z.~Kou, H.~Zheng, and C.~Xu, ``What comprises a good talking-head video generation?: A survey and benchmark,'' \emph{arXiv}, 2020.

\bibitem{chensurvey}
Z.~Chen \emph{et~al.}, ``A survey on talking head generation,'' \emph{Journal of Computer-Aided Design \& Computer Graphics}, 2023.

\bibitem{zhu2023human}
W.~Zhu, X.~Ma, D.~Ro, H.~Ci, J.~Zhang, J.~Shi, F.~Gao, Q.~Tian, and Y.~Wang, ``Human motion generation: A survey,'' \emph{IEEE Transactions on Pattern Analysis and Machine Intelligence}, 2023.

\bibitem{liao2024appearance}
F.~Liao, X.~Zou, and W.~Wong, ``Appearance and pose-guided human generation: A survey,'' \emph{ACM Computing Surveys}, vol.~56, no.~5, pp. 1--35, 2024.

\bibitem{10.1145/3575656}
T.~Sha, W.~Zhang, T.~Shen, Z.~Li, and T.~Mei, ``Deep person generation: A survey from the perspective of face, pose, and cloth synthesis,'' \emph{ACM Computing Surveys}, vol.~55, no.~12, 2023.

\bibitem{hore2010image}
A.~Hore and D.~Ziou, ``Image quality metrics: Psnr vs. ssim,'' in \emph{ICPR}, 2010.

\bibitem{wang2004image}
Z.~Wang, A.~C. Bovik, H.~R. Sheikh, and E.~P. Simoncelli, ``Image quality assessment: from error visibility to structural similarity,'' \emph{IEEE transactions on image processing}, vol.~13, no.~4, pp. 600--612, 2004.

\bibitem{zhang2018unreasonable}
R.~Zhang, P.~Isola, A.~A. Efros, E.~Shechtman, and O.~Wang, ``The unreasonable effectiveness of deep features as a perceptual metric,'' in \emph{CVPR}, 2018.

\bibitem{heusel2017gans}
M.~Heusel, H.~Ramsauer, T.~Unterthiner, B.~Nessler, and S.~Hochreiter, ``Gans trained by a two time-scale update rule converge to a local nash equilibrium,'' \emph{NeurIPS}, 2017.

\bibitem{unterthiner2018towards}
T.~Unterthiner, S.~Van~Steenkiste, K.~Kurach, R.~Marinier, M.~Michalski, and S.~Gelly, ``Towards accurate generative models of video: A new metric \& challenges,'' \emph{arXiv}, 2018.

\bibitem{mocogan2018}
S.~Tulyakov, M.-Y. Liu, X.~Yang, and J.~Kautz, ``Mocogan: Decomposing motion and content for video generation,'' in \emph{CVPR}, 2018.

\bibitem{liu2024evalcrafter}
Y.~Liu, X.~Cun, X.~Liu, X.~Wang, Y.~Zhang, H.~Chen, Y.~Liu, T.~Zeng, R.~Chan, and Y.~Shan, ``Evalcrafter: Benchmarking and evaluating large video generation models,'' in \emph{CVPR}, 2024.

\bibitem{yoon2020speech}
Y.~Yoon, B.~Cha, J.-H. Lee, M.~Jang, J.~Lee, J.~Kim, and G.~Lee, ``Speech gesture generation from the trimodal context of text, audio, and speaker identity,'' \emph{ACM Transactions on Graphics (TOG)}, vol.~39, no.~6, pp. 1--16, 2020.

\bibitem{magicpose2023}
D.~Chang, Y.~Shi, Q.~Gao, H.~Xu, J.~Fu, G.~Song, Q.~Yan, Y.~Zhu, X.~Yang, and M.~Soleymani, ``Magicpose: Realistic human poses and facial expressions retargeting with identity-aware diffusion,'' in \emph{ICML}, 2023.

\bibitem{li2021ai}
R.~Li, S.~Yang, D.~A. Ross, and A.~Kanazawa, ``Ai choreographer: Music conditioned 3d dance generation with aist++,'' in \emph{ICCV}, 2021.

\bibitem{esser2023structure}
P.~Esser, J.~Chiu, P.~Atighehchian, J.~Granskog, and A.~Germanidis, ``Structure and content-guided video synthesis with diffusion models,'' in \emph{ICCV}, 2023.

\bibitem{salimans2016improved}
T.~Salimans, I.~Goodfellow, W.~Zaremba, V.~Cheung, A.~Radford, and X.~Chen, ``Improved techniques for training gans,'' \emph{NeurIPS}, 2016.

\bibitem{liu2022learning}
X.~Liu, Q.~Wu, H.~Zhou, Y.~Xu, R.~Qian, X.~Lin, X.~Zhou, W.~Wu, B.~Dai, and B.~Zhou, ``Learning hierarchical cross-modal association for co-speech gesture generation,'' in \emph{CVPR}, 2022.

\bibitem{wu2023exploring}
H.~Wu, E.~Zhang, L.~Liao, C.~Chen, J.~Hou, A.~Wang, W.~Sun, Q.~Yan, and W.~Lin, ``Exploring video quality assessment on user generated contents from aesthetic and technical perspectives,'' in \emph{ICCV}, 2023.

\bibitem{yang2012articulated}
Y.~Yang and D.~Ramanan, ``Articulated human detection with flexible mixtures of parts,'' \emph{IEEE transactions on pattern analysis and machine intelligence}, vol.~35, no.~12, pp. 2878--2890, 2012.

\bibitem{siarohin2019first}
A.~Siarohin, S.~Lathuili{\`e}re, S.~Tulyakov, E.~Ricci, and N.~Sebe, ``First order motion model for image animation,'' \emph{NeurIPS}, 2019.

\bibitem{zhao2022thin}
J.~Zhao and H.~Zhang, ``Thin-plate spline motion model for image animation,'' in \emph{CVPR}, 2022.

\bibitem{gorelick2007actions}
L.~Gorelick, M.~Blank, E.~Shechtman, M.~Irani, and R.~Basri, ``Actions as space-time shapes,'' \emph{IEEE transactions on pattern analysis and machine intelligence}, vol.~29, no.~12, pp. 2247--2253, 2007.

\bibitem{soomro2012ucf101}
K.~Soomro, A.~R. Zamir, and M.~Shah, ``Ucf101: A dataset of 101 human actions classes from videos in the wild,'' \emph{arXiv}, 2012.

\bibitem{ionescu2013human3}
C.~Ionescu, D.~Papava, V.~Olaru, and C.~Sminchisescu, ``Human3. 6m: Large scale datasets and predictive methods for 3d human sensing in natural environments,'' \emph{IEEE transactions on pattern analysis and machine intelligence}, vol.~36, no.~7, pp. 1325--1339, 2013.

\bibitem{shahroudy2016ntu}
A.~Shahroudy, J.~Liu, T.-T. Ng, and G.~Wang, ``Ntu rgb+ d: A large scale dataset for 3d human activity analysis,'' in \emph{CVPR}, 2016.

\bibitem{sharjeel2023har}
S.~M. Rajput, M.~Bilal, and A.~Habib, ``Human activity recognition (har - video dataset),'' 2023.

\bibitem{pumarola20193dpeople}
A.~Pumarola, J.~Sanchez-Riera, G.~Choi, A.~Sanfeliu, and F.~Moreno-Noguer, ``3dpeople: Modeling the geometry of dressed humans,'' in \emph{ICCV}, 2019.

\bibitem{sadoughi2015msp}
N.~Sadoughi, Y.~Liu, and C.~Busso, ``Msp-avatar corpus: Motion capture recordings to study the role of discourse functions in the design of intelligent virtual agents,'' in \emph{FG}, 2015.

\bibitem{chan2019everybody}
C.~Chan, S.~Ginosar, T.~Zhou, and A.~A. Efros, ``Everybody dance now,'' in \emph{ICCV}, 2019.

\bibitem{TikTok2021}
Y.~Jafarian and H.~S. Park, ``Learning high fidelity depths of dressed humans by watching social media dance videos,'' in \emph{CVPR}, 2021.

\bibitem{guo2021danceit}
X.~Guo, Y.~Zhao, and J.~Li, ``Danceit: music-inspired dancing video synthesis,'' \emph{IEEE Transactions on Image Processing}, vol.~30, pp. 5559--5572, 2021.

\bibitem{disco2024}
T.~Wang, L.~Li, K.~Lin, Y.~Zhai, C.-C. Lin, Z.~Yang, H.~Zhang, Z.~Liu, and L.~Wang, ``Disco: Disentangled control for realistic human dance generation,'' in \emph{CVPR}, 2024.

\bibitem{chen2017deep}
L.~Chen, S.~Srivastava, Z.~Duan, and C.~Xu, ``Deep cross-modal audio-visual generation,'' in \emph{ACM MM Workshops}, 2017.

\bibitem{li2018creating}
B.~Li, X.~Liu, K.~Dinesh, Z.~Duan, and G.~Sharma, ``Creating a multitrack classical music performance dataset for multimodal music analysis: Challenges, insights, and applications,'' \emph{IEEE Transactions on Multimedia}, vol.~21, no.~2, pp. 522--535, 2018.

\bibitem{liu2016deepfashion}
Z.~Liu, P.~Luo, S.~Qiu, X.~Wang, and X.~Tang, ``Deepfashion: Powering robust clothes recognition and retrieval with rich annotations,'' in \emph{CVPR}, 2016.

\bibitem{dwnet2019}
P.~Zablotskaia, A.~Siarohin, B.~Zhao, and L.~Sigal, ``Dwnet: Dense warp-based network for pose-guided human video generation,'' \emph{arXiv}, 2019.

\bibitem{jiang2023text2performer}
Y.~Jiang, S.~Yang, T.~L. Koh, W.~Wu, C.~C. Loy, and Z.~Liu, ``Text2performer: Text-driven human video generation,'' in \emph{ICCV}, 2023.

\bibitem{ju2023humanart}
X.~Ju, A.~Zeng, W.~Jianan, X.~Qiang, and Z.~Lei, ``Human-art: A versatile human-centric dataset bridging natural and artificial scenes,'' in \emph{CVPR}, 2023.

\bibitem{msasl}
H.~R.~V. Joze and O.~Koller, ``Ms-asl: A large-scale data set and benchmark for understanding american sign language,'' \emph{arXiv}, 2018.

\bibitem{camgoz2018neural}
N.~C. Camgoz, S.~Hadfield, O.~Koller, H.~Ney, and R.~Bowden, ``Neural sign language translation,'' in \emph{CVPR}, 2018.

\bibitem{duarte2021how2sign}
A.~Duarte, S.~Palaskar, L.~Ventura, D.~Ghadiyaram, K.~DeHaan, F.~Metze, J.~Torres, and X.~Giro-i Nieto, ``How2sign: a large-scale multimodal dataset for continuous american sign language,'' in \emph{CVPR}, 2021.

\bibitem{luo2020arbee}
Y.~Luo, J.~Ye, R.~B. Adams, J.~Li, M.~G. Newman, and J.~Z. Wang, ``Arbee: Towards automated recognition of bodily expression of emotion in the wild,'' \emph{International journal of computer vision}, vol. 128, pp. 1--25, 2020.

\bibitem{liu2023cross}
L.~Liu and L.~Liu, ``Cross-modal mutual learning for cued speech recognition,'' in \emph{ICASSP}, 2023.

\bibitem{liu2024computation}
L.~Liu, L.~Liu, and H.~Li, ``Computation and parameter efficient multi-modal fusion transformer for cued speech recognition,'' \emph{IEEE/ACM Transactions on Audio, Speech, and Language Processing}, 2024.

\bibitem{ginosar2019learning}
S.~Ginosar, A.~Bar, G.~Kohavi, C.~Chan, A.~Owens, and J.~Malik, ``Learning individual styles of conversational gesture,'' in \emph{CVPR}, 2019.

\bibitem{liu2022audio}
X.~Liu, Q.~Wu, H.~Zhou, Y.~Du, W.~Wu, D.~Lin, and Z.~Liu, ``Audio-driven co-speech gesture video generation,'' in \emph{NeurIPS}, 2022.

\bibitem{yoonICRA19}
Y.~Yoon, W.-R. Ko, M.~Jang, J.~Lee, J.~Kim, and G.~Lee, ``Robots learn social skills: End-to-end learning of co-speech gesture generation for humanoid robots,'' in \emph{ICRA}, 2019.

\bibitem{siarohin2021motion}
A.~Siarohin, O.~J. Woodford, J.~Ren, M.~Chai, and S.~Tulyakov, ``Motion representations for articulated animation,'' in \emph{CVPR}, 2021.

\bibitem{OpenPose2017}
Z.~Cao, T.~Simon, S.-E. Wei, and Y.~Sheikh, ``Realtime multi-person 2d pose estimation using part affinity fields,'' in \emph{CVPR}, 2017.

\bibitem{yang2023effective}
Z.~Yang, A.~Zeng, C.~Yuan, and Y.~Li, ``Effective whole-body pose estimation with two-stages distillation,'' in \emph{ICCV}, 2023.

\bibitem{kendall2015posenet}
A.~Kendall, M.~Grimes, and R.~Cipolla, ``Posenet: A convolutional network for real-time 6-dof camera relocalization,'' in \emph{ICCV}, 2015.

\bibitem{wang2020deep}
J.~Wang, K.~Sun, T.~Cheng, B.~Jiang, C.~Deng, Y.~Zhao, D.~Liu, Y.~Mu, M.~Tan, X.~Wang \emph{et~al.}, ``Deep high-resolution representation learning for visual recognition,'' \emph{IEEE transactions on pattern analysis and machine intelligence}, vol.~43, no.~10, pp. 3349--3364, 2020.

\bibitem{stackpose18}
A.~Newell, K.~Yang, and J.~Deng, ``Stacked hourglass networks for human pose estimation,'' \emph{arXiv}, 2016.

\bibitem{choutas2020monocular}
V.~Choutas, G.~Pavlakos, T.~Bolkart, D.~Tzionas, and M.~J. Black, ``Monocular expressive body regression through body-driven attention,'' in \emph{ECCV}, 2020.

\bibitem{fang2022alphapose}
H.-S. Fang, J.~Li, H.~Tang, C.~Xu, H.~Zhu, Y.~Xiu, Y.-L. Li, and C.~Lu, ``Alphapose: Whole-body regional multi-person pose estimation and tracking in real-time,'' \emph{IEEE Transactions on Pattern Analysis and Machine Intelligence}, vol.~45, no.~6, pp. 7157--7173, 2022.

\bibitem{zhu2023motionbert}
W.~Zhu, X.~Ma, Z.~Liu, L.~Liu, W.~Wu, and Y.~Wang, ``Motionbert: A unified perspective on learning human motion representations,'' in \emph{ICCV}, 2023.

\bibitem{loper2023smpl}
M.~Loper, N.~Mahmood, J.~Romero, G.~Pons-Moll, and M.~J. Black, ``Smpl: A skinned multi-person linear model,'' in \emph{Seminal Graphics Papers: Pushing the Boundaries, Volume 2}, 2023, pp. 851--866.

\bibitem{pavlakos2019expressive}
G.~Pavlakos, V.~Choutas, N.~Ghorbani, T.~Bolkart, A.~A. Osman, D.~Tzionas, and M.~J. Black, ``Expressive body capture: 3d hands, face, and body from a single image,'' in \emph{CVPR}, 2019.

\bibitem{mmflow2021}
M.~Contributors, ``Mmflow: Openmmlab optical flow toolbox and benchmark,'' 2021.

\bibitem{ilg2017flownet}
E.~Ilg, N.~Mayer, T.~Saikia, M.~Keuper, A.~Dosovitskiy, and T.~Brox, ``Flownet 2.0: Evolution of optical flow estimation with deep networks,'' in \emph{CVPR}, 2017.

\bibitem{teed2020raft}
Z.~Teed and J.~Deng, ``Raft: Recurrent all-pairs field transforms for optical flow,'' in \emph{ECCV}, 2020.

\bibitem{mahjourian2018unsupervised}
R.~Mahjourian, M.~Wicke, and A.~Angelova, ``Unsupervised learning of depth and ego-motion from monocular video using 3d geometric constraints,'' in \emph{CVPR}, 2018.

\bibitem{godard2019digging}
C.~Godard, O.~Mac~Aodha, M.~Firman, and G.~J. Brostow, ``Digging into self-supervised monocular depth estimation,'' in \emph{ICCV}, 2019.

\bibitem{depthanything2024}
L.~Yang, B.~Kang, Z.~Huang, X.~Xu, J.~Feng, and H.~Zhao, ``Depth anything: Unleashing the power of large-scale unlabeled data,'' in \emph{CVPR}, 2024.

\bibitem{densepose2018}
R.~A. G{\"u}ler, N.~Neverova, and I.~Kokkinos, ``Densepose: Dense human pose estimation in the wild,'' in \emph{CVPR}, 2018.

\bibitem{rombach2022high}
R.~Rombach, A.~Blattmann, D.~Lorenz, P.~Esser, and B.~Ommer, ``High-resolution image synthesis with latent diffusion models,'' in \emph{CVPR}, 2022.

\bibitem{kim2024human}
T.~Kim, C.~Kang, J.~Park, D.~Jeong, C.~Yang, S.-J. Kang, and K.~Kong, ``Human motion aware text-to-video generation with explicit camera control,'' in \emph{WACV}, 2024.

\bibitem{stoll2020signsynth}
S.~Stoll, S.~Hadfield, and R.~Bowden, ``Signsynth: Data-driven sign language video generation,'' in \emph{ECCV}, 2020.

\bibitem{natarajan2022development}
B.~Natarajan, E.~Rajalakshmi, R.~Elakkiya, K.~Kotecha, A.~Abraham, L.~A. Gabralla, and V.~Subramaniyaswamy, ``Development of an end-to-end deep learning framework for sign language recognition, translation, and video generation,'' \emph{IEEE Access}, vol.~10, pp. 104\,358--104\,374, 2022.

\bibitem{fang2024signllm}
S.~Fang, L.~Wang, C.~Zheng, Y.~Tian, and C.~Chen, ``Signllm: Sign languages production large language models,'' \emph{arXiv}, 2024.

\bibitem{cornett1967cued}
R.~O. Cornett, ``Cued speech,'' \emph{American annals of the deaf}, pp. 3--13, 1967.

\bibitem{liu2019pilot}
L.~Liu and G.~Feng, ``A pilot study on mandarin chinese cued speech,'' \emph{American Annals of the Deaf}, vol. 164, no.~4, pp. 496--518, 2019.

\bibitem{lei2024bridge}
W.~Lei, L.~Liu, and J.~Wang, ``Bridge to non-barrier communication: Gloss-prompted fine-grained cued speech gesture generation with diffusion model,'' \emph{arXiv}, 2024.

\bibitem{wang2024dance}
X.~Wang, H.~Wang, D.~Liu, and W.~Cai, ``Dance any beat: Blending beats with visuals in dance video generation,'' \emph{arXiv}, 2024.

\bibitem{jia2023music2play}
R.~Jia and S.~Pang, ``Music2play: Audio-driven instrumental animation,'' in \emph{CAC}, 2023.

\bibitem{liao2020speech2video}
M.~Liao, S.~Zhang, P.~Wang, H.~Zhu, X.~Zuo, and R.~Yang, ``Speech2video synthesis with 3d skeleton regularization and expressive body poses,'' in \emph{ACCV}, 2020.

\bibitem{zhang2024dr2}
C.~Zhang, C.~Wang, Y.~Zhao, S.~Cheng, L.~Luo, and X.~Guo, ``Dr2: Disentangled recurrent representation learning for data-efficient speech video synthesis,'' in \emph{WACV}, 2024.

\bibitem{hogue2024diffted}
S.~Hogue, C.~Zhang, H.~Daruger, Y.~Tian, and X.~Guo, ``Diffted: One-shot audio-driven ted talk video generation with diffusion-based co-speech gestures,'' in \emph{CVPR}, 2024.

\bibitem{gao2023high}
Y.~Gao, Y.~Zhou, J.~Wang, X.~Li, X.~Ming, and Y.~Lu, ``High-fidelity and freely controllable talking head video generation,'' in \emph{CVPR}, 2023.

\bibitem{gan2023efficient}
Y.~Gan, Z.~Yang, X.~Yue, L.~Sun, and Y.~Yang, ``Efficient emotional adaptation for audio-driven talking-head generation,'' in \emph{ICCV}, 2023.

\bibitem{van2017neural}
A.~Van Den~Oord, O.~Vinyals \emph{et~al.}, ``Neural discrete representation learning,'' \emph{NeurIPS}, 2017.

\bibitem{vividpose2024}
Q.~Wang, Z.~Jiang, C.~Xu, J.~Zhang, Y.~Wang, X.~Zhang, Y.~Cao, W.~Cao, C.~Wang, and Y.~Fu, ``Vividpose: Advancing stable video diffusion for realistic human image animation,'' \emph{arXiv}, 2024.

\bibitem{posevideo2018}
C.~Yang, Z.~Wang, X.~Zhu, C.~Huang, J.~Shi, and D.~Lin, ``Pose guided human video generation,'' in \emph{ECCV}, 2018.

\bibitem{poseanimation2021}
J.~S. Yoon, L.~Liu, V.~Golyanik, K.~Sarkar, H.~S. Park, and C.~Theobalt, ``Pose-guided human animation from a single image in the wild,'' in \emph{CVPR}, 2021.

\bibitem{HumanVideo2018}
H.~Cai, C.~Bai, Y.-W. Tai, and C.-K. Tang, ``Deep video generation, prediction and completion of human action sequences,'' in \emph{ECCV}, 2018.

\bibitem{disentangled2019}
L.~Yang, Z.~Zhao, S.~Wang, S.~Wang, S.~Ma, and W.~Gao, ``Disentangled human action video generation via decoupled learning,'' in \emph{ICMEW}, 2019.

\bibitem{everydance2019}
C.~Chan, S.~Ginosar, T.~Zhou, and A.~A. Efros, ``Everybody dance now,'' in \emph{ICCV}, 2019.

\bibitem{longhuman2020}
N.~Fushishita, A.~Tejero-de Pablos, Y.~Mukuta, and T.~Harada, ``Long-term human video generation of multiple futures using poses,'' in \emph{ECCV}, 2020.

\bibitem{Twostreamvan2020}
X.~Sun, H.~Xu, and K.~Saenko, ``Twostreamvan: Improving motion modeling in video generation,'' in \emph{WACV}, 2020.

\bibitem{SGWGAN2021}
C.~Kissel, C.~K{\"u}mmel, D.~Ritter, and K.~Hildebrand, ``Pose-guided sign language video gan with dynamic lambda,'' \emph{arXiv}, 2021.

\bibitem{cgan}
M.~Mirza and S.~Osindero, ``Conditional generative adversarial nets,'' \emph{arXiv}, 2014.

\bibitem{pix2pix}
P.~Isola, J.-Y. Zhu, T.~Zhou, and A.~A. Efros, ``Image-to-image translation with conditional adversarial networks,'' in \emph{CVPR}, 2017.

\bibitem{pix2pixHD}
T.-C. Wang, M.-Y. Liu, J.-Y. Zhu, A.~Tao, J.~Kautz, and B.~Catanzaro, ``High-resolution image synthesis and semantic manipulation with conditional gans,'' in \emph{CVPR}, 2018.

\bibitem{motionfollower2024}
S.~Tu, Q.~Dai, Z.~Zhang, S.~Xie, Z.-Q. Cheng, C.~Luo, X.~Han, Z.~Wu, and Y.-G. Jiang, ``Motionfollower: Editing video motion via lightweight score-guided diffusion,'' \emph{arXiv}, 2024.

\bibitem{mimicmotion2024}
Y.~Zhang, J.~Gu, L.-W. Wang, H.~Wang, J.~Cheng, Y.~Zhu, and F.~Zou, ``Mimicmotion: High-quality human motion video generation with confidence-aware pose guidance,'' \emph{arXiv}, 2024.

\bibitem{unianimate2024}
X.~Wang, S.~Zhang, C.~Gao, J.~Wang, X.~Zhou, Y.~Zhang, L.~Yan, and N.~Sang, ``Unianimate: Taming unified video diffusion models for consistent human image animation,'' \emph{arXiv}, 2024.

\bibitem{LDMs}
R.~Rombach, A.~Blattmann, D.~Lorenz, P.~Esser, and B.~Ommer, ``High-resolution image synthesis with latent diffusion models,'' in \emph{CVPR}, 2022.

\bibitem{SVD}
A.~Blattmann, T.~Dockhorn, S.~Kulal, D.~Mendelevitch, M.~Kilian, D.~Lorenz, Y.~Levi, Z.~English, V.~Voleti, A.~Letts \emph{et~al.}, ``Stable video diffusion: Scaling latent video diffusion models to large datasets,'' \emph{arXiv}, 2023.

\bibitem{alignlatents}
A.~Blattmann, R.~Rombach, H.~Ling, T.~Dockhorn, S.~W. Kim, S.~Fidler, and K.~Kreis, ``Align your latents: High-resolution video synthesis with latent diffusion models,'' in \emph{CVPR}, 2023.

\bibitem{ControlNet}
L.~Zhang, A.~Rao, and M.~Agrawala, ``Adding conditional control to text-to-image diffusion models,'' in \emph{ICCV}, 2023.

\bibitem{DWPose2023}
Z.~Yang, A.~Zeng, C.~Yuan, and Y.~Li, ``Effective whole-body pose estimation with two-stages distillation,'' in \emph{ICCV}, 2023.

\bibitem{dreampose2023}
J.~Karras, A.~Holynski, T.-C. Wang, and I.~Kemelmacher-Shlizerman, ``Dreampose: Fashion image-to-video synthesis via stable diffusion,'' in \emph{ICCV}, 2023.

\bibitem{magicanimate2024}
Z.~Xu, J.~Zhang, J.~H. Liew, H.~Yan, J.-W. Liu, C.~Zhang, J.~Feng, and M.~Z. Shou, ``Magicanimate: Temporally consistent human image animation using diffusion model,'' in \emph{CVPR}, 2024.

\bibitem{human4dit2024}
R.~Shao, Y.~Pang, Z.~Zheng, J.~Sun, and Y.~Liu, ``Human4dit: Free-view human video generation with 4d diffusion transformer,'' \emph{arXiv}, 2024.

\bibitem{SMPL2023}
M.~Loper, N.~Mahmood, J.~Romero, G.~Pons-Moll, and M.~J. Black, ``Smpl: A skinned multi-person linear model,'' in \emph{Seminal Graphics Papers: Pushing the Boundaries, Volume 2}, 2023.

\bibitem{Latte2024}
X.~Ma, Y.~Wang, G.~Jia, X.~Chen, Z.~Liu, Y.-F. Li, C.~Chen, and Y.~Qiao, ``Latte: Latent diffusion transformer for video generation,'' \emph{arXiv}, 2024.

\bibitem{actions2007}
L.~Gorelick, M.~Blank, E.~Shechtman, M.~Irani, and R.~Basri, ``Actions as space-time shapes,'' \emph{IEEE transactions on pattern analysis and machine intelligence}, vol.~29, no.~12, pp. 2247--2253, 2007.

\bibitem{ck2010}
P.~Lucey, J.~F. Cohn, T.~Kanade, J.~Saragih, Z.~Ambadar, and I.~Matthews, ``The extended cohn-kanade dataset (ck+): A complete dataset for action unit and emotion-specified expression,'' in \emph{CVPR workshops}, 2010.

\bibitem{disentangling2024}
J.~Liu, K.~Yu, M.~Feng, X.~Guo, and M.~Cui, ``Disentangling foreground and background motion for enhanced realism in human video generation,'' \emph{arXiv}, 2024.

\bibitem{leo2023}
Y.~Wang, X.~Ma, X.~Chen, C.~Chen, A.~Dantcheva, B.~Dai, and Y.~Qiao, ``Leo: Generative latent image animator for human video synthesis,'' \emph{arXiv}, 2023.

\bibitem{LIA2022}
Y.~Wang, D.~Yang, F.~Bremond, and A.~Dantcheva, ``Latent image animator: Learning to animate images via latent space navigation,'' \emph{arXiv}, 2022.

\bibitem{TaichiHD2019}
A.~Siarohin, S.~Lathuili{\`e}re, S.~Tulyakov, E.~Ricci, and N.~Sebe, ``First order motion model for image animation,'' \emph{NeurIPS}, 2019.

\bibitem{FaceForensics2018}
A.~R{\"o}ssler, D.~Cozzolino, L.~Verdoliva, C.~Riess, J.~Thies, and M.~Nie{\ss}ner, ``Faceforensics: A large-scale video dataset for forgery detection in human faces,'' \emph{arXiv}, 2018.

\bibitem{CelebV-HQ2022}
H.~Zhu, W.~Wu, W.~Zhu, L.~Jiang, S.~Tang, L.~Zhang, Z.~Liu, and C.~C. Loy, ``Celebv-hq: A large-scale video facial attributes dataset,'' in \emph{ECCV}, 2022.

\bibitem{dreamoving2023}
M.~Feng, J.~Liu, K.~Yu, Y.~Yao, Z.~Hui, X.~Guo, X.~Lin, H.~Xue, C.~Shi, X.~Li \emph{et~al.}, ``Dreamoving: A human video generation framework based on diffusion models,'' \emph{arXiv}, 2023.

\bibitem{ZoeDepth2023}
S.~F. Bhat, R.~Birkl, D.~Wofk, P.~Wonka, and M.~M{\"u}ller, ``Zoedepth: Zero-shot transfer by combining relative and metric depth,'' \emph{arXiv}, 2023.

\bibitem{GroundedSAM2023}
S.~Liu, Z.~Zeng, T.~Ren, F.~Li, H.~Zhang, J.~Yang, C.~Li, J.~Yang, H.~Su, J.~Zhu \emph{et~al.}, ``Grounding dino: Marrying dino with grounded pre-training for open-set object detection,'' \emph{arXiv}, 2023.

\bibitem{champ2024}
S.~Zhu, J.~L. Chen, Z.~Dai, Y.~Xu, X.~Cao, Y.~Yao, H.~Zhu, and S.~Zhu, ``Champ: Controllable and consistent human image animation with 3d parametric guidance,'' \emph{arXiv}, 2024.

\bibitem{poseanimate2024}
B.~Zhu, F.~Wang, T.~Lu, P.~Liu, J.~Su, J.~Liu, Y.~Zhang, Z.~Wu, Y.-G. Jiang, and G.-J. Qi, ``Poseanimate: Zero-shot high fidelity pose controllable character animation,'' \emph{arXiv}, 2024.

\bibitem{humanmatting2020}
J.~Liu, Y.~Yao, W.~Hou, M.~Cui, X.~Xie, C.~Zhang, and X.-s. Hua, ``Boosting semantic human matting with coarse annotations,'' in \emph{CVPR}, 2020.

\bibitem{smpler-x2024}
Z.~Cai, W.~Yin, A.~Zeng, C.~Wei, Q.~Sun, W.~Yanjun, H.~E. Pang, H.~Mei, M.~Zhang, L.~Zhang \emph{et~al.}, ``Smpler-x: Scaling up expressive human pose and shape estimation,'' \emph{NeurIPS}, 2024.

\bibitem{follow-your-posev2}
J.~Xue, H.~Wang, Q.~Tian, Y.~Ma, A.~Wang, Z.~Zhao, S.~Min, W.~Zhao, K.~Zhang, H.-Y. Shum \emph{et~al.}, ``Follow-your-pose v2: Multiple-condition guided character image animation for stable pose control,'' \emph{arXiv}, 2024.

\bibitem{lu2023handrefinerrefiningmalformedhands}
W.~Lu, Y.~Xu, J.~Zhang, C.~Wang, and D.~Tao, ``Handrefiner: Refining malformed hands in generated images by diffusion-based conditional inpainting,'' \emph{arXiv}, 2023.

\end{thebibliography}


\begin{thebibliography}{1}
\bibliographystyle{IEEEtran}

\bibitem{ref1}
{\it{Mathematics Into Type}}. American Mathematical Society. [Online]. Available: https://www.ams.org/arc/styleguide/mit-2.pdf

\bibitem{ref2}
T. W. Chaundy, P. R. Barrett and C. Batey, {\it{The Printing of Mathematics}}. London, U.K., Oxford Univ. Press, 1954.

\bibitem{ref3}
F. Mittelbach and M. Goossens, {\it{The \LaTeX Companion}}, 2nd ed. Boston, MA, USA: Pearson, 2004.

\bibitem{ref4}
G. Gr\"atzer, {\it{More Math Into LaTeX}}, New York, NY, USA: Springer, 2007.

\bibitem{ref5}M. Letourneau and J. W. Sharp, {\it{AMS-StyleGuide-online.pdf,}} American Mathematical Society, Providence, RI, USA, [Online]. Available: http://www.ams.org/arc/styleguide/index.html

\bibitem{ref6}
H. Sira-Ramirez, ``On the sliding mode control of nonlinear systems,'' \textit{Syst. Control Lett.}, vol. 19, pp. 303--312, 1992.

\bibitem{ref7}
A. Levant, ``Exact differentiation of signals with unbounded higher derivatives,''  in \textit{Proc. 45th IEEE Conf. Decis.
Control}, San Diego, CA, USA, 2006, pp. 5585--5590. DOI: 10.1109/CDC.2006.377165.

\bibitem{ref8}
M. Fliess, C. Join, and H. Sira-Ramirez, ``Non-linear estimation is easy,'' \textit{Int. J. Model., Ident. Control}, vol. 4, no. 1, pp. 12--27, 2008.

\bibitem{ref9}
R. Ortega, A. Astolfi, G. Bastin, and H. Rodriguez, ``Stabilization of food-chain systems using a port-controlled Hamiltonian description,'' in \textit{Proc. Amer. Control Conf.}, Chicago, IL, USA,
2000, pp. 2245--2249.

\end{thebibliography}


 




\vfill

\end{document}